\documentclass[journal,10pt,letterpaper]{IEEEtran}
\usepackage[T1]{fontenc}
\usepackage{microtype}
\usepackage{amsmath,amssymb}
\usepackage{graphicx}
\usepackage{array,multirow}
\usepackage{arydshln}
\usepackage{placeins}
\usepackage{xcolor}
\usepackage{cite}
\usepackage[hidelinks]{hyperref}

\title{Learning Accurate End-Effector Trajectory Tracking with Inertial Feedback for a Quadruped Manipulator}
\author{Dohyeon~Kim\textsuperscript{1}, Seungyong~Han\textsuperscript{2,3,*}, and Yeongtae~Jung\textsuperscript{2,3,*}
\thanks{\textsuperscript{1}Graduate School of Mechanical-Aerospace-Electric Convergence Engineering, Jeonbuk National University, Jeollabuk-do 54896, Republic of Korea.}
\thanks{\textsuperscript{2}Department of Mechanical System Engineering, Jeonbuk National University, Jeollabuk-do 54896, Republic of Korea.}
\thanks{\textsuperscript{3}Advanced Transportation Machinery Research Center, Jeonbuk National University, Jeollabuk-do 54896, Republic of Korea.}
\thanks{\textsuperscript{*}Corresponding authors: Seungyong Han (\href{mailto:hansy@jbnu.ac.kr}{hansy@jbnu.ac.kr}) and Yeongtae Jung (\href{mailto:ytjung@jbnu.ac.kr}{ytjung@jbnu.ac.kr}).}
}

\hypersetup{
  pdftitle={Learning Accurate End-Effector Trajectory Tracking with Inertial Feedback for a Quadruped Manipulator},
  pdfauthor={Dohyeon Kim, Seungyong Han, Yeongtae Jung}
}

\begin{document}
\pagestyle{plain}
\raggedbottom
\maketitle
\thispagestyle{plain}

\begin{abstract}
Accurate end-effector trajectory tracking on quadruped manipulators is challenging because uncertainties in actuators, foot-ground contact, and geometric errors are propagated through the leg-arm kinematic chain. Detecting such errors requires sensing beyond joint encoders that can capture the effects of these uncertainties on the robot's motion, since joint encoders only provide measurements at the actuated joints and cannot fully observe mechanical errors such as backlash, dimensional inaccuracies, or structural deformation.
This work addresses these challenges using a reinforcement learning framework with end-effector (EE) inertial feedback, targeting local high-accuracy manipulation while maintaining a nominal body pose and support configuration without intentional body motion or foot repositioning.
The learned policy uses EE inertial measurements to compensate for tracking errors induced by uncertainties throughout the leg-arm system that cannot be fully inferred from proprioception alone.
Simulation and real-world experiments on continuous EE trajectory tracking show overall improvements relative to nominal and proprioception-only baselines.
In real-world experiments, the proposed policy reduces the MoCap-based RMSE by 17.3\% and 28.0\% relative to the nominal baseline under the two in-distribution (ID) command conditions, respectively, and by an additional 12.6\% and 15.9\% relative to the proprioception-only baseline.
Furthermore, the trajectory-wise RMSE between FK-reconstructed and motion-capture-measured end-effector positions averaged $12.61 \pm 2.15$~mm (mean $\pm$ standard deviation) across 24 controller--trajectory combinations, revealing a substantial discrepancy between nominal kinematic reconstruction and measured end-effector motion.
\end{abstract}


\section{Introduction}

\IEEEPARstart{L}{egged} manipulators have been widely studied for manipulation in unstructured environments where fixed-base industrial arms and wheeled mobile manipulators are limited by accessibility and terrain constraints.
Quadruped manipulators~\cite{decoupled1, wholebody_RL2, wholebody_RL3, wholebody_RL4, wholebody_RL5, wholebody_RL6, wholebody_RL7, wholebody_modelbased1, wholebody_modelbased2, wholebody_modelbased3, wholebody_modelbased4, wheeled_wholebody1, wheeled_wholebody2} are particularly attractive because four-legged support provides inherent stability and terrain adaptability.
However, manipulation control remains difficult due to the strong dynamic coupling between the legged base and the arm.
Prior studies have investigated quadruped manipulation for point-reaching tasks~\cite{wholebody_RL2,wholebody_RL3,wholebody_modelbased3}, force-controlled interaction~\cite{wholebody_RL1}, and gross manipulation behaviors~\cite{wholebody_RL4,wholebody_RL5}.
Several of these approaches exploit coordinated leg, base, and arm motions to extend the reachable workspace or accomplish whole-body manipulation tasks.

Despite significant progress in quadruped manipulation~\cite{wholebody_RL1, wholebody_RL2, wholebody_RL3, wholebody_RL4, wholebody_RL5, wholebody_RL6, wholebody_RL7}, accurate end-effector (EE) trajectory tracking remains challenging because hardware uncertainties propagate through the leg-arm system.
Whole-body motions for manipulation, including intentional body motion, foot repositioning, and locomotion, can alter the foot-ground contact geometry and effective contact impedance, thereby introducing additional uncertainty into the realized base pose.
Therefore, when workspace expansion is not required, maintaining a nominal body pose and support configuration is advantageous for accurate trajectory tracking because it avoids these additional variations, although at the cost of a limited reachable workspace.
Accordingly, this work focuses on achieving accurate continuous EE position tracking without intentional body motion or foot repositioning.

Even without intentional body motion, a floating-base quadruped manipulator is supported through compliant and contact-dependent foot-ground interactions rather than being rigidly fixed to the world.  As a result, foot and ground compliance, slip, wear of foot materials, and contact configuration changes introduce significant base-motion uncertainties that directly affect the end-effector.
In addition to these floating-base uncertainties, actuator uncertainties, backlash, structural compliance, manufacturing and assembly tolerances, payload-induced deformation, and kinematic calibration errors propagate through the long leg-arm chain and accumulate at the end-effector.
These uncertainties are difficult to model accurately or infer from joint encoders alone, and even small deviations in the base or intermediate links can result in non-negligible task-space tracking errors.
This challenge is further amplified by the low gear-reduction actuation commonly used in legged robots, such as quasi-direct-drive (QDD) actuators~\cite{QDD1}, which often employ low-reduction-ratio planetary gearboxes with non-negligible backlash that can contribute to end-effector tracking errors.

Among prior studies on quadruped manipulation, relatively few have considered continuous end-effector trajectory tracking~\cite{wholebody_RL4, decoupled1, wholebody_modelbased1}, and their evaluations remain limited.
Existing works either do not provide direct external ground-truth measurements of the end-effector~\cite{decoupled1,wholebody_RL4}, or evaluate end-effector tracking primarily through forward kinematics (FK) calculation based on measured arm joint states while employing motion-capture measurements only for estimating the robot base state~\cite{wholebody_modelbased1}. These approaches rely on kinematic reconstruction of the end-effector state from joint measurements and estimated or measured base states rather than direct task-space observations.
As a result, modeling errors that accumulate along the leg-arm chain may remain only partially observable.
Achieving high end-effector trajectory-tracking accuracy therefore requires either accurate compensation of these unmodeled effects or additional sensing capable of directly capturing the realized end-effector motion in task space.
This requirement is particularly difficult to satisfy in high-DOF floating-base systems, where accumulated end-effector errors cannot be fully inferred from joint-space measurements alone.

To improve task-space accuracy beyond joint-space proprioception, prior studies have employed exteroceptive sensing systems such as motion-capture systems, cameras~\cite{global_feedback2, wholebody_RL4, wholebody_RL5}, total stations~\cite{totalstation1}, and laser trackers~\cite{laser_tracker1}.
While these systems can provide task-space observations or global pose information, their deployment in mobile manipulation is often limited by cost, infrastructure requirements, occlusion, sensing latency, update rate, and environmental sensitivity.

Kindle et al.~\mbox{\cite{IMU1}} used two high-accuracy accelerometers mounted on the chassis base and at the manipulator base, both upstream of the manipulator arm. Their factor graph combined these measurements with deflection and backlash models and forward kinematics to improve end-effector position estimates. The accelerometer measurement model accounts for gravity and sensor bias, constraining gravity-referenced roll and pitch without directly observing the effects of arm-side errors at the end effector. However, the inertial measurements were used only for end-effector position estimation; neither compensation control nor improved trajectory tracking was demonstrated.

Stadelmann et al.~\cite{IMU4} fused iGPS and inertial measurement unit (IMU) measurements to provide smooth, high-rate, and low-latency end-effector pose estimates for end-effector pose correction in large-scale fixed-base robotic systems.
However, the approach remains dependent on a dedicated external tracking system for pose estimation and correction.

Other studies focused on improving end-effector state estimation without demonstrating a corresponding improvement in hardware closed-loop trajectory tracking.
Cvitanic et al.~\mbox{\cite{IMU3}} fused end-effector acceleration and angular velocity with pose observations, using Kalman filtering for motions without rotation and particle filtering for motions with rotation. Their simulations used laser-tracker pose measurements, whereas their hardware experiments fused robot-encoder data with IMU measurements and evaluated improvements in velocity estimation.
Korayem et al.~\mbox{\cite{IMU5}} mounted an IMU on the end effector of a cable-driven parallel robot and fused its linear acceleration and angular velocity measurements with motor-encoder measurements using a Kalman-filter-based method to estimate end-effector position and orientation in simulation and a hardware experiment. However, the IMU measurements were not used as feedback for the hardware controller, which relied on motor-encoder measurements and forward kinematics.
The hardware evaluations in these studies were limited to state estimation; the fused estimates were not applied to reduce end-effector trajectory-tracking errors. Achieving stable feedback control using these estimates for tracking compensation can be challenging, particularly in the presence of IMU measurement noise.

Sandy and Buchli~\cite{IMU6} and Hashemi and Mattila~\cite{IMU7} combined end-effector inertial and visual sensing for end-effector state estimation and task-space control.
Even at the state-estimation level, the reported real-world position RMSEs were approximately 50-90~mm in their respective estimation experiments. Such centimeter-scale estimation errors are substantial relative to the millimeter-scale tracking-error reductions sought in precision manipulation.
In~\cite{IMU6}, the real-world closed-loop end-effector tracking accuracy was not quantitatively evaluated using an independent external ground-truth measurement system.
In~\cite{IMU7}, the real-world experiment evaluated the proposed observer, whereas the observer-based closed-loop controller was evaluated only in simulation.

While gravity-referenced roll and pitch can be used for end-effector orientation feedback control, achieving accurate continuous end-effector position tracking using inertial feedback remains challenging.
We use end-effector inertial observations to generate learned residual arm commands for continuous end-effector position tracking on a floating-base quadruped manipulator, without external pose feedback during tracking.
Unlike joint encoders, which only provide measurements at the actuated joints, the EE IMU provides local inertial information reflecting the realized end-effector motion after the effects of system uncertainties have propagated through the leg-arm system, providing complementary motion information that cannot be fully inferred from joint-space measurements alone.
The effectiveness of the inertial feedback is evaluated by its effect on the final control objective---the realized end-effector tracking error---rather than by state-estimation accuracy alone.
Leveraging this property, the proposed framework applies residual learning~\cite{Residual_RL1} to generate corrective arm commands on top of the nominal kinematics-based controller.
The residual policy corrects the nominal arm commands while retaining the existing leg controller.

However, exploiting direct end-effector inertial feedback in reinforcement learning introduces significant sim-to-real challenges.
IMU signals are commonly low-pass filtered to suppress measurement noise, introducing phase lag that degrades rapid feedback correction.
Moreover, IMU observations in the learning simulator are derived from idealized robot dynamics and fail to capture real-world hardware effects, including sensor noise, transmission ripple, gear meshing, backlash, structural deflection, and actuator dynamics.
We therefore employ a two-stage alignment procedure.
First, joint-space chirp data are used to identify low-order closed-loop arm dynamics by matching the simulated and real joint responses through the Covariance Matrix Adaptation Evolution Strategy (CMA-ES) algorithm.
A neural residual model then corrects the remaining discrepancy between simulated and real end-effector inertial observations. Without the proposed alignment, the policy using EE inertial feedback exhibited severe oscillatory behavior during hardware deployment, precluding further hardware testing.

The proposed feedback formulation is evaluated through both simulation and real-world continuous trajectory tracking.
In the hardware experiments, an independent motion-capture system is used for initialization and performance evaluation; its measurements are not provided to the controller during trajectory tracking.
Unlike evaluations based on kinematic reconstruction without independent end-effector ground truth~\mbox{\cite{decoupled1,wholebody_RL4,wholebody_modelbased1}}, we directly quantify the discrepancy between nominal FK reconstruction and independently measured end-effector motion during continuous trajectory tracking in a fixed reference frame. This frame is established from the initial base pose and remains fixed throughout each trial, preserving the contribution of subsequent base motion to the measured end-effector trajectory.
The comparison between residual policies with and without end-effector inertial feedback directly evaluates whether the additional inertial feedback reduces the realized closed-loop tracking error, while the FK-MoCap comparison reveals tracking-error components that are not captured by joint-space proprioception and nominal kinematic reconstruction alone.
To the best of our knowledge, this is the first work to incorporate end-effector inertial observations, including the projected gravity vector, angular velocity, and linear acceleration, as additional feedback in a residual-learning framework for continuous end-effector position tracking on a quadruped manipulator, without relying on external sensing infrastructure for feedback during trajectory tracking.
The primary contributions of this work are summarized as follows:
\begin{itemize}
    \item \emph{End-effector inertial residual learning:}
    
A residual arm controller that combines joint-space proprioception with end-effector inertial measurements to compensate for motion effects of leg--arm uncertainties during continuous tracking, while retaining the existing leg controller and requiring no online external pose feedback.

    \item \emph{Two-stage sim-to-real alignment:}
    A two-stage sim-to-real alignment procedure that identifies low-order closed-loop arm actuator dynamics and models the remaining discrepancy in end-effector inertial observations using a neural residual network, enabling hardware deployment of the inertial-feedback policy.

    \item \emph{Continuous trajectory-tracking evaluation:}
    
Simulation and hardware evaluations demonstrating improved accuracy in continuous end-effector position tracking, with the realized hardware tracking errors independently measured using motion capture.

    \item \emph{Real-world uncertainty analysis:}
    A quantitative comparison of nominal FK reconstruction and independent MoCap measurements in a fixed reference frame during continuous trajectory tracking on a quadruped manipulator. In contrast to FK-based tracking evaluations~\mbox{\cite{decoupled1,wholebody_modelbased1}}, this analysis reveals discrepancies in realized end-effector motion that include the effects of base motion and arm-side uncertainties.
\end{itemize}

\section{Problem Statement and Formulation}

\subsection{Problem Statement}
Accurate end-effector tracking on a quadruped manipulator fundamentally differs from that of a fixed-base manipulator because the robot operates on a floating base. In a quadruped manipulator, the realized EE motion is influenced not only by the commanded arm motion but also by contact-dependent base motion, actuator dynamics, backlash, structural compliance, kinematic errors, and other system uncertainties. These uncertainties propagate through the floating-base leg-arm kinematic chain and accumulate at the end effector, resulting in task-space tracking errors.

A fundamental challenge is that these propagated errors cannot be fully observed using joint encoders alone. Joint encoders measure only the actuated joints and therefore cannot directly capture the realized end-effector motion after uncertainty has propagated through the leg-arm system. Consequently, controllers relying solely on joint-space proprioception have limited ability to compensate for these errors. The objective of this work is therefore to reduce the propagated end-effector tracking errors without relying on external sensing infrastructure.

\subsection{Problem Formulation}
End-effector tracking of a quadruped manipulator is formulated as a reinforcement learning problem because the propagated tracking errors arise from multiple coupled uncertainties that are difficult to model explicitly. The learned policy outputs residual arm corrections, and the problem is modeled as a Markov decision process, defined by the tuple $\mathcal{M}
=
\langle
\mathcal{S},
\mathcal{A},
\mathcal{P},
\mathcal{R},
\gamma,
\mu
\rangle ,$

where $\mathcal{S}$ and $\mathcal{A}$ denote the state and action spaces, and $\mathcal{P}: \mathcal{S} \times \mathcal{A} \rightarrow \mathcal{S}$ represents the state transition function induced by the robot dynamics, the nominal controllers, and the low-level joint servo controllers.
The reward function $\mathcal{R}: \mathcal{S} \times \mathcal{A} \rightarrow \mathbb{R}$ evaluates tracking performance and regularization terms, $\gamma \in (0,1)$ is the discount factor, and $\mu$ denotes the initial state distribution.
At each time step, the policy receives the observation and outputs a residual arm action to maximize the expected discounted return.

\section{Method}
\label{sec:method}

\begin{figure*}[!t]
    \centering
    \includegraphics[width=\linewidth]{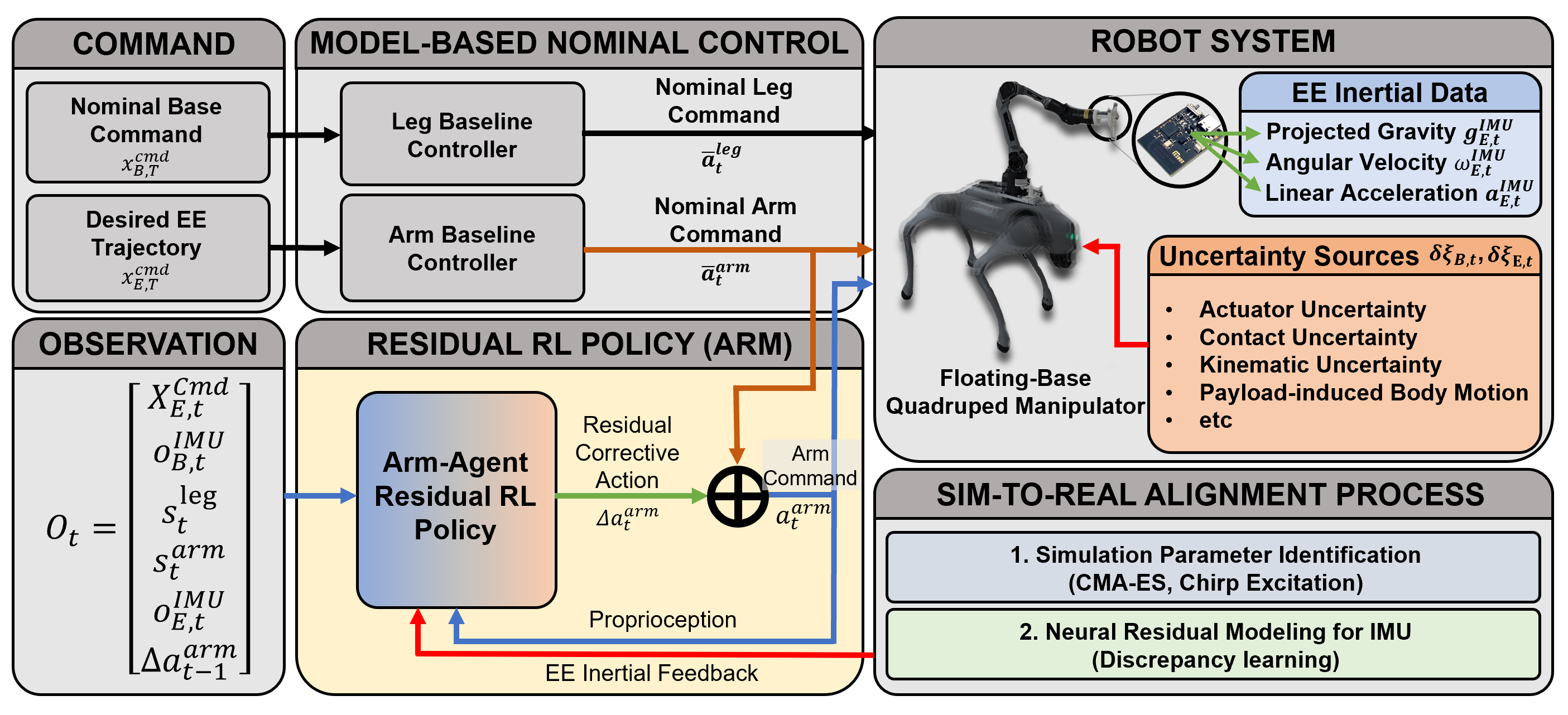}
    \caption{Overview of the proposed control framework. Nominal leg and arm controllers generate baseline commands, while an arm-agent residual RL policy uses proprioceptive and end-effector inertial feedback to compensate for propagated tracking errors. Sim-to-real alignment combines simulation parameter identification and neural residual modeling of IMU discrepancies.}
    \label{fig:controller_architecture}
\end{figure*}

\subsection{Task-Space Error Propagation Formulation}

This section formulates how uncertainties in the leg--arm system propagate to the end effector and explains why joint-space proprioception alone is insufficient for compensating the resulting task-space errors.
Let $\mathbf{T}^{\mathrm{nom}}_{WB,t} \in SE(3)$ denote the nominal base pose with respect to the world frame, generated by the kinematics-based leg controller.
The actual base pose $\mathbf{T}_{WB,t} \in SE(3)$ on the real robot generally deviates from this nominal pose due to contact-dependent foot-ground interactions, actuator and kinematic uncertainties, and payload-induced body motion.
This deviation is expressed as
\begin{equation}
    \mathbf{T}_{WB,t}
    =
    \mathbf{T}^{\mathrm{nom}}_{WB,t}
    \exp\left( \delta \boldsymbol{\xi}_{B,t}^{\wedge} \right)
\end{equation}
where $\delta \boldsymbol{\xi}_{B,t} \in \mathbb{R}^{6}$ is the base perturbation represented in the Lie algebra of $SE(3)$, and $(\cdot)^{\wedge}$ denotes the wedge operator.
The base perturbation is induced by multiple sources of uncertainty:
\begin{equation}
    \delta \boldsymbol{\xi}_{B,t}
    =
    d_B
    \left(
    \Delta \mathbf{k}_{\mathrm{leg}},
    \Delta \mathbf{c}_{\mathrm{contact}},
    \Delta \mathbf{a}_{\mathrm{actuator}},
    \mathbf{w}_t
    \right)
\end{equation}
where $\Delta \mathbf{k}_{\mathrm{leg}}$ denotes leg-side kinematic uncertainties, $\Delta \mathbf{c}_{\mathrm{contact}}$ represents contact uncertainties such as foot slip and ground interaction, $\Delta \mathbf{a}_{\mathrm{actuator}}$ denotes actuator uncertainty, and $\mathbf{w}_t$ denotes other uncertainties not explicitly represented in the formulation.
Therefore, accurately eliminating $\delta \boldsymbol{\xi}_{B,t}$ would require precise modeling and system identification of the entire robot, including contact-dependent base dynamics, actuator behavior, and payload-dependent body motion.

The nominal end-effector pose can be written as
\begin{equation}
    \mathbf{T}^{\mathrm{nom}}_{WE,t}
    =
    \mathbf{T}^{\mathrm{nom}}_{WB,t}
    \mathbf{T}_{BE}
    \left(
    \mathbf{q}^{\mathrm{arm}}_t;
    \boldsymbol{\theta}_{\mathrm{arm}}
    \right)
\end{equation}
where $\mathbf{q}^{\mathrm{arm}}_t \in \mathbb{R}^{6}$ is the arm joint configuration, $\boldsymbol{\theta}_{\mathrm{arm}}$ denotes the nominal arm kinematic parameters, and $\mathbf{T}_{BE}(\cdot)$ represents the FK from the base frame to the end-effector frame.
In practice, the actual end-effector pose is affected by both base perturbation and arm-side modeling errors:
\begin{equation}
    \mathbf{T}_{WE,t}
    =
    \mathbf{T}_{WB,t}
    \mathbf{T}_{BE}
    \left(
    \mathbf{q}^{\mathrm{arm}}_t + \delta \mathbf{q}^{\mathrm{arm}}_t;
    \boldsymbol{\theta}_{\mathrm{arm}} + \Delta \boldsymbol{\theta}_{\mathrm{arm}}
    \right)
\end{equation}
where $\delta \mathbf{q}^{\mathrm{arm}}_t$ denotes arm actuation and joint-level tracking errors, and $\Delta \boldsymbol{\theta}_{\mathrm{arm}}$ represents arm-side modeling uncertainties, including kinematic calibration errors, backlash, and compliance.

For small deviations, the resulting end-effector pose error can be approximated as
\begin{equation}
    \delta \boldsymbol{\xi}_{E,t}
    \approx
    \mathbf{J}_{B,t} \delta \boldsymbol{\xi}_{B,t}
    +
    \mathbf{J}_{A,t} \delta \mathbf{q}^{\mathrm{arm}}_t
    +
    \mathbf{J}_{\theta,t} \Delta \boldsymbol{\theta}_{\mathrm{arm}}
\end{equation}
where $\mathbf{J}_{B,t}$ maps base-pose deviations to end-effector pose errors, $\mathbf{J}_{A,t}$ is the arm Jacobian, and $\mathbf{J}_{\theta,t}$ describes the sensitivity of the end-effector pose to arm-side modeling uncertainties.
These effects are transmitted through the leg--arm kinematic chain and jointly contribute to the resulting task-space tracking error. Since these propagated effects are not directly observable from joint encoder measurements alone, the resulting end-effector tracking error cannot be fully inferred from joint-space proprioception. Consequently, compensating for the propagated task-space errors requires observations beyond joint-space proprioception.

\subsection{Residual Reinforcement Learning with Inertial Feedback}
To address this limitation, a residual reinforcement learning framework is proposed that uses an end-effector-mounted inertial measurement unit (IMU) as additional feedback beyond joint-space proprioception.
The end-effector-mounted IMU provides inertial measurements of the realized end-effector motion after uncertainty effects have propagated through the leg--arm system. Unlike joint encoders, which provide only local joint-state measurements, the IMU captures end-effector-level motion information that cannot be fully inferred from joint-space proprioception alone.

The proposed controller combines nominal kinematics-based leg and arm controllers with a learned residual arm policy, as shown in Figure~\ref{fig:controller_architecture}. Rather than explicitly modeling and suppressing uncertainty-induced base and arm-side errors, the residual policy learns corrective arm commands that compensate for their effect on the realized end-effector motion. In this architecture, the nominal leg command is generated as
\begin{equation}
    \bar{\mathbf{a}}^{\mathrm{leg}}_t
    =
    f^{\mathrm{leg}}_{\mathrm{IK}}
    \left(
    \mathbf{x}^{\mathrm{cmd}}_{B,t}
    \right)
\end{equation}
where $\mathbf{x}^{\mathrm{cmd}}_{B,t}$ is the commanded base pose.
The nominal arm command is generated by the arm inverse-kinematics controller:
\begin{equation}
    \bar{\mathbf{a}}^{\mathrm{arm}}_t
    =
    f^{\mathrm{arm}}_{\mathrm{IK}}
    \left(
    \mathbf{x}^{\mathrm{cmd}}_{E,t},
    \mathbf{q}^{\mathrm{arm}}_t
    \right)
\end{equation}
where
$\bar{\mathbf{a}}^{\mathrm{arm}}_t \in \mathbb{R}^{6}$
denotes the nominal arm joint command.
Using the proprioceptive and end-effector inertial observations, the residual policy generates a corrective arm command
$\Delta\mathbf{a}^{\mathrm{arm}}_t$.
The final arm command is given by
\begin{equation}
    \mathbf{a}^{\mathrm{arm}}_t
    =
    \bar{\mathbf{a}}^{\mathrm{arm}}_t
    +
    \Delta \mathbf{a}^{\mathrm{arm}}_t
\end{equation}
The complete command applied to the robot is
\begin{equation}
    \mathbf{a}_t
    =
    \left[
    \bar{\mathbf{a}}^{\mathrm{leg}}_t,
    \mathbf{a}^{\mathrm{arm}}_t
    \right]
\end{equation}
The resulting joint position commands are tracked by low-level joint-space proportional-derivative (PD) controllers that operate at $200~\mathrm{Hz}$, while the residual policy is updated at $50~\mathrm{Hz}$.

\subsection{Observation and Action Spaces}

At time step $t$, the policy observation is designed to combine the commanded end-effector pose, the base inertial state, leg and arm proprioception, end-effector inertial measurements, and the previous residual arm action so that the residual policy can compensate for propagated task-space errors.
The robot proprioception includes the base, leg, and arm states.
The base inertial observation is defined as
\begin{equation}
    \mathbf{o}^{\mathrm{IMU}}_{B,t}
    =
    \left[
    \boldsymbol{\omega}^{\mathrm{IMU}}_{B,t},
    \mathbf{g}^{\mathrm{IMU}}_{B,t}
    \right],
\end{equation}
where $\boldsymbol{\omega}^{\mathrm{IMU}}_{B,t}$ and $\mathbf{g}^{\mathrm{IMU}}_{B,t}$ denote the base angular velocity and projected gravity vector measured by the onboard base IMU, respectively.

The leg and arm states are defined as
\begin{equation}
    \mathbf{s}^{\mathrm{leg}}_t
    =
    \left[
    \mathbf{q}^{\mathrm{leg}}_t,
    \dot{\mathbf{q}}^{\mathrm{leg}}_t
    \right],
    \qquad
    \mathbf{s}^{\mathrm{arm}}_t
    =
    \left[
    \mathbf{q}^{\mathrm{arm}}_t,
    \dot{\mathbf{q}}^{\mathrm{arm}}_t
    \right]
\end{equation}
Although the policy does not command the legs, leg proprioception provides information on the supporting configuration and contact-induced motion affecting the floating base and EE. The leg controller is intentionally kept fixed so that the residual policy compensates for propagated end-effector errors without requiring the leg controller itself to be learned or the entire leg-base system to be precisely aligned between simulation and hardware.

To augment joint-space proprioception with end-effector-level inertial feedback, the policy uses the projected gravity vector, angular velocity, and linear acceleration.
The projected gravity vector provides gravity-referenced roll and pitch information, whereas the yaw angle is omitted because it is not observable from gravity and is susceptible to heading drift.
Angular velocity and linear acceleration capture the rotational and translational inertial components of the realized end-effector motion, including the effects of uncertainties that have propagated through the leg--arm system but are not fully observable from joint-space proprioception alone.
These quantities are also selected because they provide complementary rotational and translational motion information without introducing additional errors associated with integration or differentiation.

The IMU observation is defined as
\begin{equation}
    \mathbf{o}^{\mathrm{IMU}}_{E,t}
    =
    \left[
    \mathbf{g}^{\mathrm{IMU}}_{E,t},
    \boldsymbol{\omega}^{\mathrm{IMU}}_{E,t},
    \mathbf{a}^{\mathrm{IMU}}_{E,t}
    \right]
\end{equation}
where
$\mathbf{g}^{\mathrm{IMU}}_{E,t}$,
$\boldsymbol{\omega}^{\mathrm{IMU}}_{E,t}$,
and
$\mathbf{a}^{\mathrm{IMU}}_{E,t}$
denote the end-effector projected gravity vector, angular velocity, and linear acceleration, respectively.
The complete policy observation is
\begin{equation}
    \mathbf{o}_t
    =
    \left[
    \mathbf{x}^{\mathrm{cmd}}_{E,t},
    \mathbf{o}^{\mathrm{IMU}}_{B,t},
    \mathbf{s}^{\mathrm{leg}}_t,
    \mathbf{s}^{\mathrm{arm}}_t,
    \mathbf{o}^{\mathrm{IMU}}_{E,t},
    \Delta \mathbf{a}^{\mathrm{arm}}_{t-1}
    \right].
\end{equation}
The previous residual action $\Delta \mathbf{a}^{\mathrm{arm}}_{t-1}$ is included to provide short-term action history.

The policy outputs a six-dimensional residual arm command:
\begin{equation}
    \Delta \mathbf{a}^{\mathrm{arm}}_t
    =
    \pi_{\boldsymbol{\theta}}
    \left(
    \mathbf{o}_t
    \right),
    \qquad
    \Delta \mathbf{a}^{\mathrm{arm}}_t
    \in
    \mathbb{R}^{6}
\end{equation}
The action represents a residual joint-position correction added to the nominal arm command, while the nominal leg command remains unchanged.

\subsection{Trajectory Command Generation}

Whereas several prior studies consider goal-point reaching~\mbox{\cite{wholebody_RL2,wholebody_RL3,wholebody_modelbased3}}, the objective of this work is continuous end-effector trajectory tracking, for which the entire commanded trajectory must remain kinematically feasible.
Candidate end-effector waypoints are sampled within a bounded workspace defined by $x \in [0.25, 0.45]$~m, $y \in [-0.10, 0.10]$~m, and $z \in [0.25, 0.35]$~m, and connected by Cartesian line segments time-parameterized with trapezoidal velocity profiles. Because Cartesian workspace constraints alone do not guarantee joint-space feasibility for the arm, inverse kinematics is evaluated at every sampled trajectory point, and candidates containing IK failures or joint-limit violations are discarded. This offline filtering process yields 1000 continuous, IK-feasible trajectories for policy training.
The workspace is intentionally bounded to represent a local region for high-accuracy manipulation and to maintain continuous IK-feasible trajectories over randomized multi-segment commands. The bounded region is not intended to characterize the overall reachable workspace. If broader workspace coverage is required, it can be achieved in several ways; for example, the precomputed feasible trajectory set can be rotated about the arm's first joint axis to cover surrounding task regions. Intentional body-pose motions for workspace expansion are avoided because even small changes in the leg-ground configuration can alter the foot-ground contact geometry and effective contact impedance, introducing additional base-pose uncertainty. Despite the bounded spatial workspace, the dynamic difficulty of the tracking task is varied through the commanded end-effector velocity and acceleration.

The training trajectories use maximum velocities of
$V_{\max}\in[0.05,\,0.10]~\mathrm{m/s}$
and maximum accelerations of
$A_{\max}\in[0.5,\,3.0]~\mathrm{m/s^2}$.
For out-of-distribution (OOD) evaluation, the same workspace and generation procedure are retained, while the velocity limit is increased to $V_{\max} \in [0.20, 0.30]$~m/s and/or the acceleration limit is increased to $A_{\max} \in [4.0, 8.0]$~m/s$^2$.
Velocity-OOD, acceleration-OOD, and combined velocity/acceleration-OOD cases are used to evaluate robustness to commands beyond the training distribution.

\subsection{Reward Design}

The reward is designed to train the residual arm policy for continuous end-effector trajectory tracking while the nominal leg controller remains fixed.
The corresponding tracking rewards are defined as
\begin{equation}
    r^{\mathrm{pos}}_{E,t}
    =
    \exp
    \left(
    -\frac{e^{\mathrm{pos}}_{E,t}}{\sigma_{\mathrm{pos}}}
    \right),
    \qquad
    r^{\mathrm{ori}}_{E,t}
    =
    \exp
    \left(
    -\frac{e^{\mathrm{ori}}_{E,t}}{\sigma_{\mathrm{ori}}}
    \right)
\end{equation}
where $e^{\mathrm{pos}}_{E,t}$ is the Euclidean distance between the desired and current end-effector positions, $e^{\mathrm{ori}}_{E,t}$ is the quaternion error magnitude between the desired and current end-effector orientations, and $\sigma_{\mathrm{pos}}$ and $\sigma_{\mathrm{ori}}$ determine the sensitivity of the corresponding reward terms.
The end-effector pose-tracking reward is defined as
\begin{equation}
    r^{\mathrm{pose}}_{E,t}
    =
    w_{\mathrm{pos}} r^{\mathrm{pos}}_{E,t}
    +
    w_{\mathrm{ori}} r^{\mathrm{ori}}_{E,t}
\end{equation}
where $w_{\mathrm{pos}}$ and $w_{\mathrm{ori}}$ denote the weighting coefficients for the position and orientation rewards, respectively.
The orientation reward encourages the end effector to maintain the desired orientation during position tracking. In the evaluations, the desired orientation is held constant, and performance is assessed primarily in terms of position-tracking accuracy.
To discourage excessive and abrupt residual corrections, the total reward is defined as
\begin{equation}
    r_t
    =
    r^{\mathrm{pose}}_{E,t}
    +
    w_{\mathrm{act}}
    \left\|
    \Delta \mathbf{a}^{\mathrm{arm}}_t
    \right\|^2
    +
    w_{\mathrm{smooth}}
    \left\|
    \Delta \mathbf{a}^{\mathrm{arm}}_t
    -
    \Delta \mathbf{a}^{\mathrm{arm}}_{t-1}
    \right\|^2
\end{equation}

The action-magnitude penalty limits deviation from the nominal controller, while the action-rate penalty promotes smooth residual commands.
The reward parameters are set to $w_{\mathrm{pos}}=50$, $w_{\mathrm{ori}}=30$, $w_{\mathrm{act}}=-0.3$, $w_{\mathrm{smooth}}=-0.3$, $\sigma_{\mathrm{pos}}=0.03$, and $\sigma_{\mathrm{ori}}=0.15$. The same reward function and parameters are used for both residual-learning policies throughout training.

\subsection{Sim-to-Real Alignment of Arm Dynamics and End-Effector Inertial Observations}

\begin{figure*}[!t]
    \centering
    \includegraphics[width=0.95\linewidth]{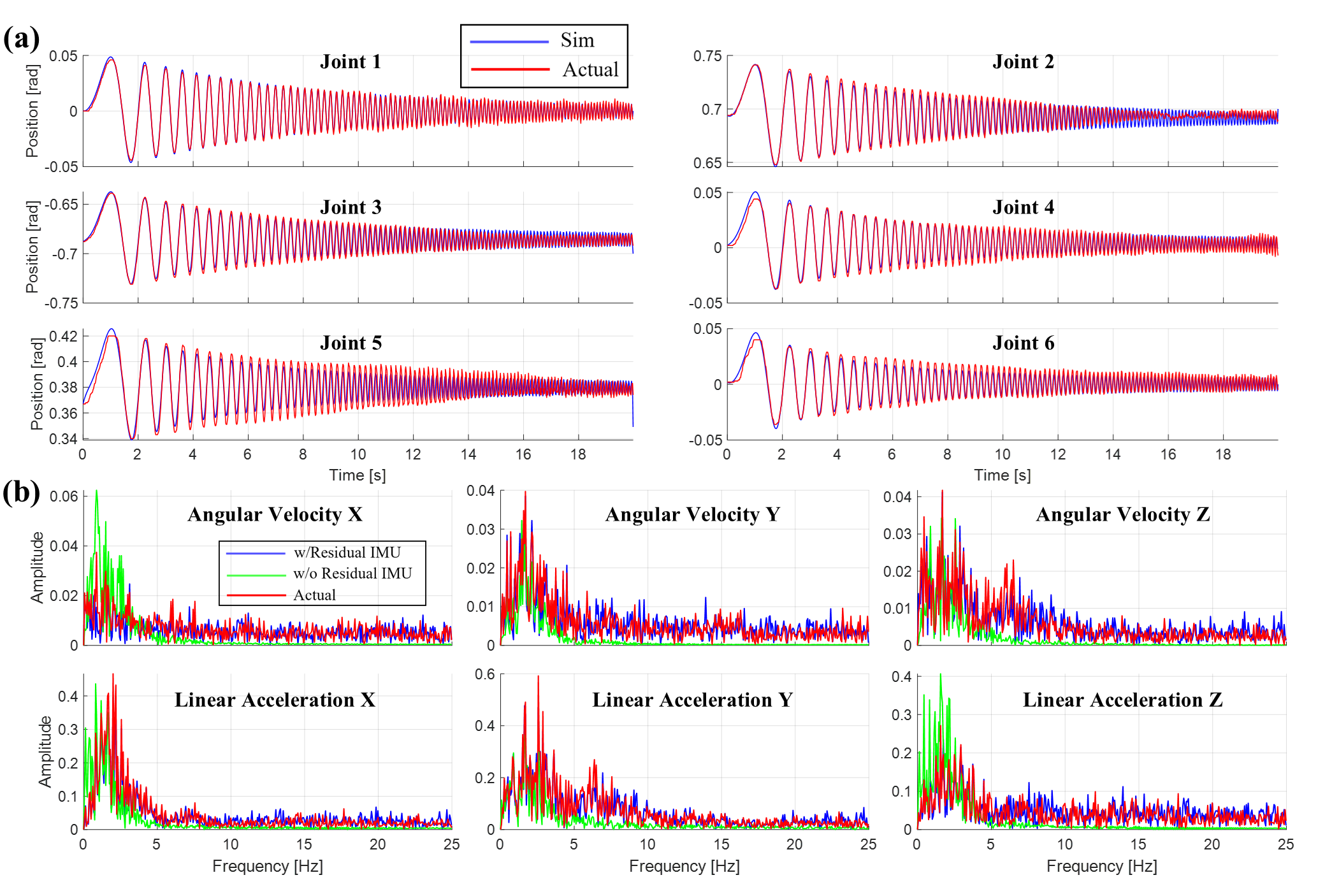}
    \caption{Sim-to-real alignment results for the closed-loop arm dynamics and end-effector IMU measurements. (a) Joint-space trajectory matching after simulation parameter identification using chirp excitation. (b) Frequency-domain comparison of end-effector IMU signals before and after applying the neural residual model.}
    \label{fig:Sysid}
\end{figure*}

Reducing the sim-to-real discrepancy in arm dynamics and end-effector IMU measurements is necessary for hardware deployment of the proposed inertial-feedback policy. To address these discrepancies, we employ a two-stage sim-to-real alignment procedure. The first stage identifies low-order closed-loop arm dynamics, while the second uses a neural residual model to compensate for the remaining end-effector inertial discrepancy. The second stage is not required for the w/o EE IMU policy because this policy does not use EE inertial observations. In preliminary hardware tests, the w/o EE IMU policy could be deployed only after the first-stage alignment, whereas the w/ EE IMU policy required both stages. In both cases, omitting the required alignment resulted in excessively large and oscillatory arm responses, precluding further hardware testing.
\subsubsection{Arm Closed-Loop Dynamics Identification}

To identify the actuator-related simulation parameters of the serial manipulator, real-world excitation data are collected by applying joint-space chirp commands to the arm. The chirp excitation sweeps the input frequency from $0.1~\mathrm{Hz}$ to $10~\mathrm{Hz}$.
Let $\boldsymbol{\phi}$ denote the set of simulation parameters to be identified, comprising the joint stiffness and damping coefficients. These parameters are selected because they strongly influence the dominant low-order closed-loop response of the arm. The identification objective is to minimize the time-domain discrepancy between the real and simulated joint trajectories:
\begin{equation}
    \boldsymbol{\phi}^{*}
    =
    \arg\min_{\boldsymbol{\phi}}
    \sum_{t=0}^{T}
    \left\|
    \mathbf{q}^{\mathrm{real}}_t
    -
    \mathbf{q}^{\mathrm{sim}}_t(\boldsymbol{\phi})
    \right\|^2 .
\end{equation}
The simulation parameters are optimized using CMA-ES to match the simulated joint trajectories to the real responses under the same chirp command.
As shown in Figure~\ref{fig:Sysid}(a), the identified simulator reproduces the dominant low-order actuator dynamics observed in the real arm. Quantitatively, the identified simulator achieves an average joint-trajectory RMSE of 0.004664 rad across the six arm joints under the chirp excitation, indicating close agreement between the identified simulation and real joint responses.

\subsubsection{Residual Modeling of End-Effector IMU Discrepancy}

Even after joint-level alignment, simulated and real IMU signals can still differ because stiffness and damping parameters alone cannot capture the remaining nonlinear sim-to-real discrepancies, as shown in Figure~\ref{fig:Sysid}(b).
Potential contributors to this discrepancy include sensor noise, transmission ripple, cogging torque, gear-mesh excitation, backlash-induced transients, structural vibration, sensor filtering, and other unmodeled hardware effects.
A neural residual model is therefore introduced to correct the remaining real-to-simulation discrepancy in the simulated end-effector angular velocity and linear acceleration.

Let $\mathbf{y}^{\mathrm{IMU,sim}}_t$ denote the simulated IMU observation consisting of the end-effector angular velocity and linear acceleration. The corrected simulated observation is defined as
\begin{equation}
\tilde{\mathbf{y}}^{\mathrm{IMU,sim}}_t
=
\mathbf{y}^{\mathrm{IMU,sim}}_t
+
f_{\psi}
\left(
\mathbf{q}^{\mathrm{arm}}_t,
\dot{\mathbf{q}}^{\mathrm{arm}}_t,
\mathbf{y}^{\mathrm{IMU,sim}}_t
\right),
\end{equation}
where $f_{\psi}$ predicts the residual IMU discrepancy from the arm joint states and the simulated inertial observation. The neural residual model uses a three-sample input history and consists of three temporal convolutional layers with 64 channels each, followed by a multilayer perceptron with three hidden layers of 512 units each. Using paired real and simulated trajectories generated under the same excitation commands, the network parameters are optimized as
\begin{equation}
\psi^{*}
=
\arg\min_{\psi}
\sum_t
\left\|
\mathbf{y}^{\mathrm{IMU,real}}_t
-
\tilde{\mathbf{y}}^{\mathrm{IMU,sim}}_t
\right\|^2 .
\end{equation}
The trained residual model is then fixed and applied to the simulated IMU observations during policy training.
The normalized frequency-domain discrepancy is defined as the L2 difference between the real and simulated one-sided FFT amplitude spectra, normalized by the real-spectrum L2 norm over all three axes. The residual model reduces this discrepancy by 46.0\% for angular velocity (0.7821 to 0.4220) and 53.3\% for linear acceleration (0.6407 to 0.2992).

\section{Simulation Analysis}

\subsection{Simulation Setup}

The proposed framework is validated using Isaac Lab~\cite{isaaclab}, a GPU-accelerated physics simulation environment.
Training is conducted using proximal policy optimization (PPO)~\cite{PPO} on an NVIDIA RTX 4070 GPU with 4,096 parallel environments.
The residual policies are implemented as multilayer perceptrons with three hidden layers of 512 units each.
To improve robustness to sim-to-real discrepancies, domain randomization is applied to observation noise, backlash, and leg-joint damping during training. Zero-mean Gaussian noise is added to the joint-velocity, EE angular-velocity, and EE linear-acceleration observations with standard deviations of 0.1~rad/s, 0.1~rad/s, and 1.0~m/s$^2$, respectively. At each reset, joint backlash is randomized within $[0.1^\circ, 0.3^\circ]$, and leg-joint damping gains are randomized within $[0.5, 1.0]$~N$\cdot$m$\cdot$s/rad.

Three control configurations are compared. The \emph{nominal} controller uses the kinematics-based controller without residual learning.
The \emph{w/o EE IMU} policy uses residual learning with proprioceptive observations only, whereas the proposed \emph{w/ EE IMU} policy additionally incorporates end-effector inertial feedback.
The evaluation includes both in-distribution (ID) commands and out-of-distribution (OOD) commands with velocities and/or accelerations beyond the training ranges. The evaluation trajectories contain 17 to 60 Cartesian line segments, with an average of $36.5 \pm 17.7$ segments per trajectory. In both simulation and hardware evaluations, the desired end-effector orientation is held constant while the end effector follows the commanded position trajectories. This setting is used to reduce the risk of motion-capture marker occlusion in the subsequent hardware experiments, since occlusion can introduce large measurement errors.

\subsection{Simulation Results}

Table~\ref{tab:sim_results} summarizes the simulated end-effector position-tracking errors under ID and OOD command conditions.
The residual policies consistently outperform the kinematics-based nominal controller, and the proposed w/ EE IMU policy achieves the lowest RMSE and MAE in all eight cases.
For the representative ID trajectory in Case~1, the w/ EE IMU policy reduces the RMSE by 46.4\% relative to the nominal controller and by 16.8\% relative to the w/o EE IMU policy.
In Cases~5 and~6, the proposed policy provides additional RMSE reductions of 11.1\% and 12.0\%, respectively, over the w/o EE IMU policy.
The performance difference between the two residual policies decreases under high-velocity OOD commands, where tracking is increasingly limited by actuator dynamics and the relatively low policy update rate.
Although the additional gain over the w/o EE IMU policy is modest in some cases with velocities and accelerations beyond the training range, the w/ EE IMU policy consistently achieves the lowest RMSE and MAE across all evaluated conditions. Based on the mean RMSE across all eight simulation cases, the proposed policy achieves a 30.5\% reduction relative to the nominal controller and an additional 5.5\% reduction relative to the w/o EE IMU policy. These results demonstrate that end-effector inertial feedback provides complementary information for compensating errors that cannot be fully inferred from proprioception alone.

\begin{table*}[!t]
\centering
\caption{Simulation Results: End-Effector Position Tracking Error}
\label{tab:sim_results}
\renewcommand{\arraystretch}{1.03}
\setlength{\tabcolsep}{2.2pt}
\small
\resizebox{\linewidth}{!}{
\begin{tabular}{cclcccccc}
\hline
\textbf{Case}
& \textbf{Cmd.}
& \textbf{Method}
& \multicolumn{3}{c}{\textbf{RMSE norm (mm)}}
& \multicolumn{3}{c}{\textbf{MAE norm (mm)}} \\
\cline{4-6} \cline{7-9}
& (Vmax, Amax)
&
& \textbf{Val.}
& \textbf{vs Nom.} (\%)
& \textbf{vs w/o} (\%)
& \textbf{Val.}
& \textbf{vs Nom.} (\%)
& \textbf{vs w/o} (\%) \\
\hline

\multirow{3}{*}{1}
& \multirow{3}{*}{\shortstack{ID\\(0.05, 0.5)}}
& Nominal & 6.34 & -- & -- & 5.83 & -- & -- \\
& & w/o EE IMU & 4.09 & 35.5 & -- & 3.82 & 34.5 & -- \\
& & w/ EE IMU  & \textbf{3.40} & \textbf{46.4} & \textbf{16.8} & \textbf{3.20} & \textbf{45.1} & \textbf{16.2} \\
\hline

\multirow{3}{*}{2}
& \multirow{3}{*}{\shortstack{ID\\(0.10, 3.0)}}
& Nominal & 8.05 & -- & -- & 7.70 & -- & -- \\
& & w/o EE IMU & 6.83 & 15.2 & -- & 6.60 & 14.3 & -- \\
& & w/ EE IMU  & \textbf{6.30} & \textbf{21.8} & \textbf{7.7} & \textbf{6.06} & \textbf{21.3} & \textbf{8.2} \\
\hline

\multirow{3}{*}{3}
& \multirow{3}{*}{\shortstack{Vel. OOD\\(0.20, 0.5)}}
& Nominal & 13.55 & -- & -- & 12.36 & -- & -- \\
& & w/o EE IMU & 8.96 & 33.9 & -- & 8.05 & 34.8 & -- \\
& & w/ EE IMU  & \textbf{8.60} & \textbf{36.5} & \textbf{4.0} & \textbf{7.54} & \textbf{39.0} & \textbf{6.3} \\
\hline

\multirow{3}{*}{4}
& \multirow{3}{*}{\shortstack{Vel. OOD\\(0.30, 3.0)}}
& Nominal & 27.95 & -- & -- & 25.26 & -- & -- \\
& & w/o EE IMU & 20.62 & 26.2 & -- & 18.45 & 27.0 & -- \\
& & w/ EE IMU  & \textbf{20.22} & \textbf{27.7} & \textbf{2.0} & \textbf{18.06} & \textbf{28.5} & \textbf{2.1} \\
\hline

\multirow{3}{*}{5}
& \multirow{3}{*}{\shortstack{Acc. OOD\\(0.05, 4.0)}}
& Nominal & 8.83 & -- & -- & 8.19 & -- & -- \\
& & w/o EE IMU & 4.30 & 51.3 & -- & 4.08 & 50.2 & -- \\
& & w/ EE IMU  & \textbf{3.82} & \textbf{56.7} & \textbf{11.1} & \textbf{3.59} & \textbf{56.1} & \textbf{12.0} \\
\hline

\multirow{3}{*}{6}
& \multirow{3}{*}{\shortstack{Acc. OOD\\(0.10, 8.0)}}
& Nominal & 8.52 & -- & -- & 8.13 & -- & -- \\
& & w/o EE IMU & 7.12 & 16.4 & -- & 6.92 & 14.8 & -- \\
& & w/ EE IMU  & \textbf{6.26} & \textbf{26.5} & \textbf{12.0} & \textbf{6.08} & \textbf{25.2} & \textbf{12.1} \\
\hline

\multirow{3}{*}{7}
& \multirow{3}{*}{\shortstack{Vel./Acc. OOD\\(0.20, 4.0)}}
& Nominal & 16.40 & -- & -- & 15.42 & -- & -- \\
& & w/o EE IMU & 14.04 & 14.4 & -- & 13.45 & 12.8 & -- \\
& & w/ EE IMU  & \textbf{13.19} & \textbf{19.6} & \textbf{6.1} & \textbf{12.66} & \textbf{17.9} & \textbf{5.8} \\
\hline

\multirow{3}{*}{8}
& \multirow{3}{*}{\shortstack{Vel./Acc. OOD\\(0.30, 8.0)}}
& Nominal & 29.34 & -- & -- & 26.06 & -- & -- \\
& & w/o EE IMU & 21.50 & 26.7 & -- & 20.17 & 22.6 & -- \\
& & w/ EE IMU  & \textbf{20.88} & \textbf{28.8} & \textbf{2.9} & \textbf{19.67} & \textbf{24.5} & \textbf{2.5} \\
\hline

\end{tabular}
}
\end{table*}

\section{Experimental Analysis}

\subsection{Experiment Setup}

\begin{figure}[!htb]
    \centering
    \includegraphics[width=0.76\linewidth]{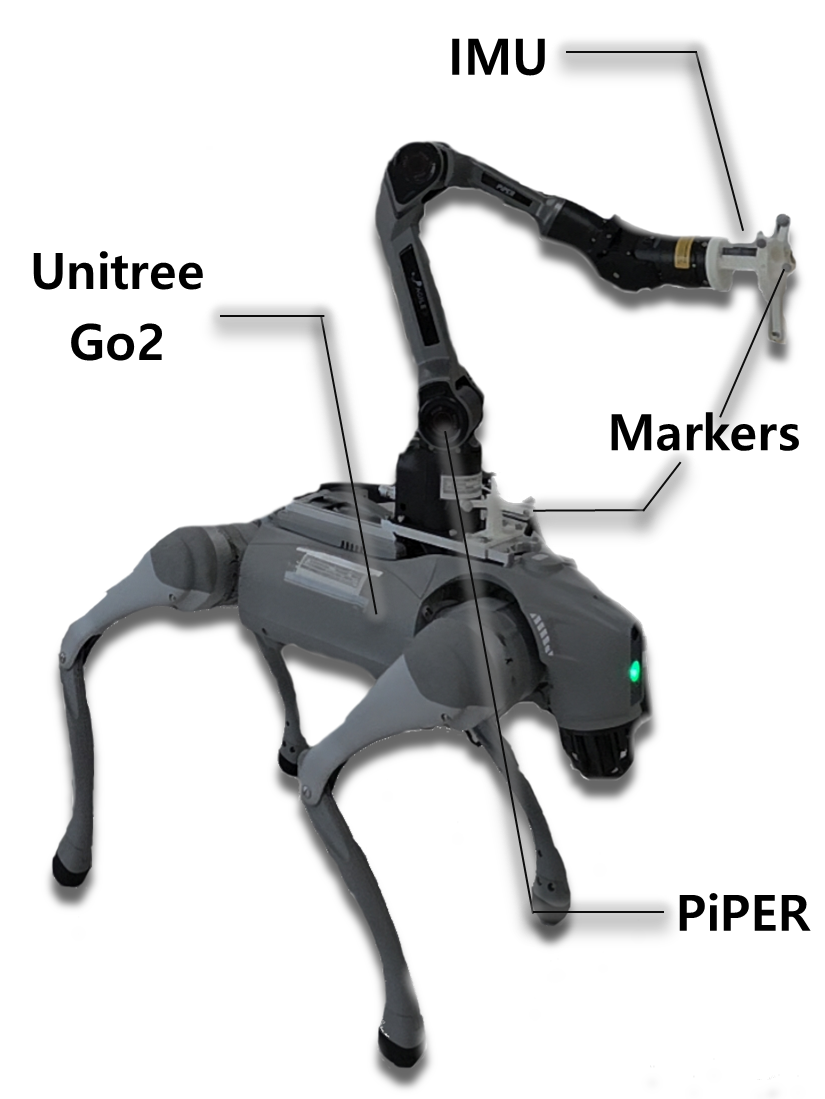}
    \caption{Experimental setup consisting of a Unitree Go2 quadruped, a mounted PiPER arm, an end-effector-mounted IMU, and motion-capture markers on the end-effector and the Go2 base.}
    \label{fig:experimental_setup}
\end{figure}

The experimental platform is an 18-DOF quadruped manipulator system composed of a 12-DOF Unitree Go2 quadruped and a 6-DOF PiPER robotic arm from AgileX Robotics, as shown in Figure~\ref{fig:experimental_setup}.
An EBMotion V6 inertial measurement unit (IMU) from E2BOX is mounted at the end effector to provide direct end-effector-level inertial feedback.
All policies are trained in simulation and directly deployed to the physical robot without additional fine-tuning.
An OptiTrack motion-capture system consisting of eight Prime 13 cameras operating at $240~\mathrm{Hz}$ is used for initialization and performance evaluation. During trajectory tracking, the data are streamed and logged in real time via UDP without being provided to the controller.
In Table~\ref{tab:exp_results} and Figure~\ref{fig:timeseries_experiment}, FK denotes the end-effector position reconstructed from forward kinematics using the measured joint states, whereas MoCap denotes the end-effector position obtained from motion-capture markers attached to the base and end effector through coordinate-frame transformation. The marker-to-model transformations were defined using CAD geometry.
The quadruped base pose is measured by the motion-capture system using four markers rigidly attached to the base. This measurement is used only during initialization to define a fixed base reference frame for each trial and for offline performance evaluation, and is not provided to either the nominal controller or the residual policy during trajectory tracking. The commanded, FK-reconstructed, and MoCap-measured end-effector positions are all expressed and evaluated in this common base frame.
The command conditions and controller configurations used in simulation are retained for the real-world experiments. Before each experimental trial, the robot is first driven to the same nominal home configuration using the nominal IK controller with low-level joint-space PD control and is allowed to settle. The measured base pose after settling is then used to define the fixed reference frame and desired base pose for trajectory tracking. Because residual leg-joint tracking errors remain under load, a brief post-initialization base transient occurs as end-effector trajectory tracking begins. The same initialization procedure is used for all controller configurations.

\subsection{Experimental Results}

Table~\ref{tab:exp_results} summarizes the real-world end-effector position-tracking errors under the same ID and OOD command conditions used in simulation.

\begin{table*}[!t]
\centering
\caption{Experimental Results: End-Effector Position Tracking Error}
\label{tab:exp_results}
\renewcommand{\arraystretch}{1.03}
\setlength{\tabcolsep}{1.7pt}
\setlength{\dashlinedash}{1.0pt}
\setlength{\dashlinegap}{1.2pt}
\small
\resizebox{\linewidth}{!}{
\begin{tabular}{cclcccccccc}
\hline
\textbf{Case}
& \textbf{Cmd.}
& \textbf{Method}
& \multicolumn{4}{c}{\textbf{RMSE norm (mm)}}
& \multicolumn{4}{c}{\textbf{MAE norm (mm)}} \\
\cline{4-7} \cline{8-11}
& (Vmax, Amax)
&
& \multicolumn{1}{:c:}{\textbf{FK}} & \textbf{MoCap}
& \textbf{vs Nom} (\%)
& \textbf{vs w/o} (\%)
& \multicolumn{1}{:c:}{\textbf{FK}} & \textbf{MoCap}
& \textbf{vs Nom} (\%)
& \textbf{vs w/o} (\%) \\
\hline

\multirow{3}{*}{1}
& \multirow{3}{*}{\shortstack{ID\\(0.05, 0.5)}}
& Nominal & \multicolumn{1}{:c:}{6.67} & 13.92 & -- & -- & \multicolumn{1}{:c:}{6.15} & 13.29 & -- & -- \\
& & w/o EE IMU & \multicolumn{1}{:c:}{6.40} & 13.18 & 5.3 & -- & \multicolumn{1}{:c:}{6.06} & 12.57 & 5.4 & -- \\
& & w/ EE IMU  & \multicolumn{1}{:c:}{5.86} & \textbf{11.51} & \textbf{17.3} & \textbf{12.6} & \multicolumn{1}{:c:}{5.53} & \textbf{10.83} & \textbf{18.5} & \textbf{13.8} \\
\hline

\multirow{3}{*}{2}
& \multirow{3}{*}{\shortstack{ID\\(0.10, 3.0)}}
& Nominal & \multicolumn{1}{:c:}{11.58} & 15.06 & -- & -- & \multicolumn{1}{:c:}{10.83} & 13.62 & -- & -- \\
& & w/o EE IMU & \multicolumn{1}{:c:}{10.23} & 12.90 & 14.4 & -- & \multicolumn{1}{:c:}{9.80} & 11.84 & 13.1 & -- \\
& & w/ EE IMU  & \multicolumn{1}{:c:}{7.94} & \textbf{10.84} & \textbf{28.0} & \textbf{15.9} & \multicolumn{1}{:c:}{7.46} & \textbf{10.10} & \textbf{25.8} & \textbf{14.7} \\
\hline

\multirow{3}{*}{3}
& \multirow{3}{*}{\shortstack{Vel. OOD\\(0.20, 0.5)}}
& Nominal & \multicolumn{1}{:c:}{12.86} & 17.46 & -- & -- & \multicolumn{1}{:c:}{11.47} & 16.14 & -- & -- \\
& & w/o EE IMU & \multicolumn{1}{:c:}{12.22} & 16.84 & 3.5 & -- & \multicolumn{1}{:c:}{11.07} & 15.56 & 3.6 & -- \\
& & w/ EE IMU  & \multicolumn{1}{:c:}{9.74} & \textbf{12.98} & \textbf{25.6} & \textbf{22.9} & \multicolumn{1}{:c:}{8.50} & \textbf{12.09} & \textbf{25.1} & \textbf{22.3} \\
\hline

\multirow{3}{*}{4}
& \multirow{3}{*}{\shortstack{Vel. OOD\\(0.30, 3.0)}}
& Nominal & \multicolumn{1}{:c:}{21.49} & 24.84 & -- & -- & \multicolumn{1}{:c:}{20.24} & 23.03 & -- & -- \\
& & w/o EE IMU & \multicolumn{1}{:c:}{20.31} & \textbf{23.47} & \textbf{5.5} & -- & \multicolumn{1}{:c:}{18.95} & \textbf{21.67} & \textbf{5.9} & -- \\
& & w/ EE IMU  & \multicolumn{1}{:c:}{19.85} & 23.60 & 5.0 & -0.6 & \multicolumn{1}{:c:}{18.35} & 21.78 & 5.5 & -0.5 \\
\hline

\multirow{3}{*}{5}
& \multirow{3}{*}{\shortstack{Acc. OOD\\(0.05, 4.0)}}
& Nominal & \multicolumn{1}{:c:}{8.13} & 15.08 & -- & -- & \multicolumn{1}{:c:}{7.05} & 14.15 & -- & -- \\
& & w/o EE IMU & \multicolumn{1}{:c:}{8.63} & 13.02 & 13.6 & -- & \multicolumn{1}{:c:}{7.24} & 12.04 & 14.9 & -- \\
& & w/ EE IMU  & \multicolumn{1}{:c:}{6.51} & \textbf{10.44} & \textbf{30.8} & \textbf{19.8} & \multicolumn{1}{:c:}{6.09} & \textbf{9.70} & \textbf{31.4} & \textbf{19.5} \\
\hline

\multirow{3}{*}{6}
& \multirow{3}{*}{\shortstack{Acc. OOD\\(0.10, 8.0)}}
& Nominal & \multicolumn{1}{:c:}{12.40} & 18.35 & -- & -- & \multicolumn{1}{:c:}{11.53} & 16.91 & -- & -- \\
& & w/o EE IMU & \multicolumn{1}{:c:}{12.18} & 15.88 & 13.5 & -- & \multicolumn{1}{:c:}{11.05} & 14.22 & 15.9 & -- \\
& & w/ EE IMU  & \multicolumn{1}{:c:}{9.88} & \textbf{12.56} & \textbf{31.5} & \textbf{20.9} & \multicolumn{1}{:c:}{9.62} & \textbf{11.42} & \textbf{32.5} & \textbf{19.7} \\
\hline

\multirow{3}{*}{7}
& \multirow{3}{*}{\shortstack{Vel./Acc. OOD\\(0.20, 4.0)}}
& Nominal & \multicolumn{1}{:c:}{18.43} & 25.89 & -- & -- & \multicolumn{1}{:c:}{17.58} & 23.92 & -- & -- \\
& & w/o EE IMU & \multicolumn{1}{:c:}{17.39} & 22.05 & 14.8 & -- & \multicolumn{1}{:c:}{16.53} & 20.41 & 14.7 & -- \\
& & w/ EE IMU  & \multicolumn{1}{:c:}{17.48} & \textbf{21.72} & \textbf{16.1} & \textbf{1.5} & \multicolumn{1}{:c:}{16.40} & \textbf{19.96} & \textbf{16.6} & \textbf{2.2} \\
\hline

\multirow{3}{*}{8}
& \multirow{3}{*}{\shortstack{Vel./Acc. OOD\\(0.30, 8.0)}}
& Nominal & \multicolumn{1}{:c:}{27.46} & 30.41 & -- & -- & \multicolumn{1}{:c:}{26.25} & 28.55 & -- & -- \\
& & w/o EE IMU & \multicolumn{1}{:c:}{26.35} & 28.76 & 5.4 & -- & \multicolumn{1}{:c:}{25.25} & 27.05 & 5.2 & -- \\
& & w/ EE IMU  & \multicolumn{1}{:c:}{25.96} & \textbf{27.73} & \textbf{8.8} & \textbf{3.6} & \multicolumn{1}{:c:}{24.83} & \textbf{26.30} & \textbf{7.9} & \textbf{2.8} \\
\hline
\end{tabular}
}
\end{table*}

\begin{figure*}[!t]
    \centering
    \begingroup
\pdfximage width \linewidth {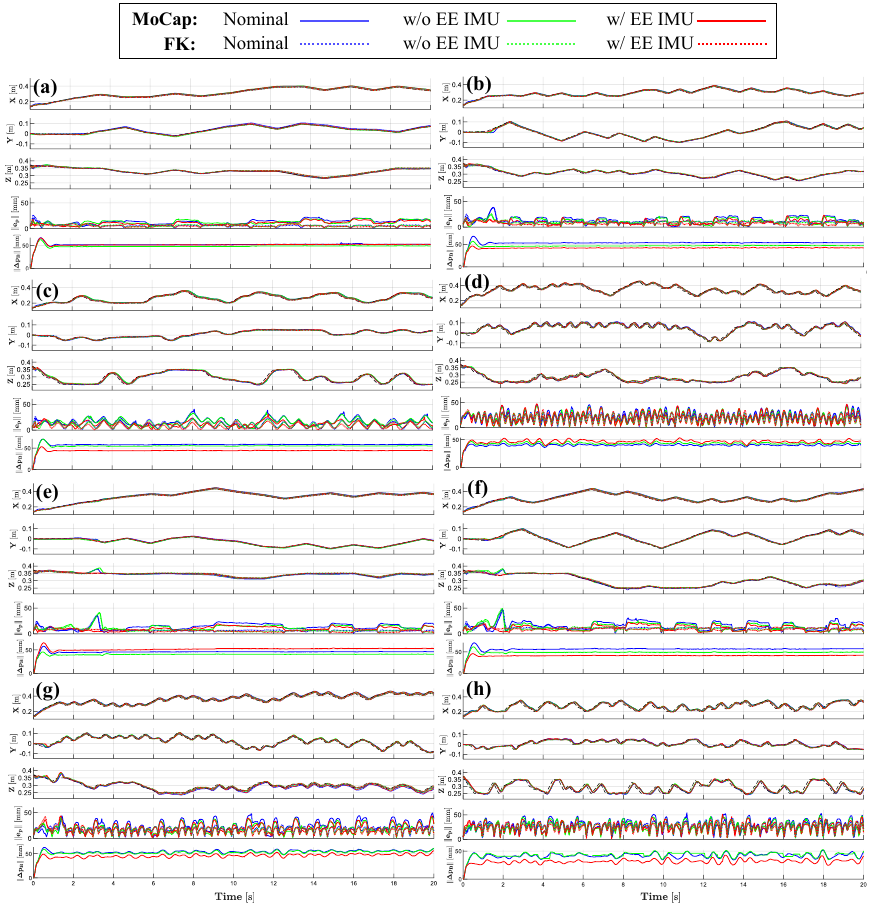}
\pdfrefximage\pdflastximage
\endgroup
    \caption{Experimental end-effector trajectory-tracking results over the complete trajectories for all eight evaluation cases. (a)-(h) Cases 1-8, respectively. For each case, the first three rows show the desired and measured Cartesian EE positions, the fourth row shows the corresponding EE position-error norms for the nominal controller, the w/o EE IMU policy, and the proposed w/ EE IMU policy, and the bottom row shows the MoCap-measured floating-base translational deviations from the reference base position for the same three controller configurations. Solid lines denote motion-capture-based measurements, dotted lines denote FK-based reconstructions, and black dashed lines denote the desired trajectories.}
    \label{fig:timeseries_experiment}
\end{figure*}

Figure~\ref{fig:timeseries_experiment} presents the complete trajectory-tracking results for all eight experimental cases, showing the measured Cartesian trajectories, corresponding position-error norms, and MoCap-measured floating-base deviations for the three control configurations. A supplementary video presents the whole-robot behavior of the proposed w/ EE IMU policy for representative ID and OOD cases, qualitatively illustrating the trajectory difficulty and corresponding leg and floating-base motions. The supplementary video also shows a brief post-initialization base transient at the onset of trajectory tracking, which is common to all controller configurations and does not persist thereafter.

Although the absolute tracking errors are larger than in simulation due to residual sim-to-real discrepancies, both residual policies outperform the nominal controller in terms of MoCap-based RMSE and MAE across all eight cases.
For the ID commands, the w/ EE IMU policy reduces the MoCap-based RMSE by 17.3\% and 28.0\% relative to the nominal controller in Cases~1 and~2, respectively.
Compared with the w/o EE IMU policy, the corresponding additional RMSE reductions are 12.6\% and 15.9\%.
The benefit of end-effector inertial feedback is also evident under acceleration-OOD commands.
In Cases~5 and~6, the proposed policy provides additional MoCap-based RMSE reductions of 19.8\% and 20.9\%, respectively, over the w/o EE IMU policy.
Under the highest-velocity and combined velocity/acceleration commands beyond the training range, the proposed w/ EE IMU policy continues to outperform the nominal controller, while showing similar or slightly better performance than the w/o EE IMU policy, consistent with the reduced performance gap between the two residual policies observed in simulation.
This suggests that tracking under aggressive motion is increasingly constrained by hardware bandwidth and the finite policy update rate.

\subsection{Experimental Analysis}

The comparison between the FK-based and MoCap-based errors provides additional insight into the source of the real-world tracking error. The FK-based error is reconstructed from the measured joint states and nominal kinematics, and therefore represents a joint-space-proprioception-based kinematic estimate rather than a direct measurement of the realized end-effector motion. In contrast, the MoCap-based error is computed from externally measured end-effector positions and reflects the combined effects of base motion, structural compliance, backlash, calibration errors, contact effects, and other hardware uncertainties.
Unlike kinematics-based tracking evaluations without independent end-effector ground truth~\mbox{\cite{wholebody_RL4,decoupled1,wholebody_modelbased1}}, the present analysis directly compares nominal FK reconstruction with independently measured end-effector positions throughout continuous trajectories in a reference frame fixed at initialization. Thus, the FK--MoCap discrepancy includes the effects of subsequent base motion as well as arm-side uncertainties.

Across the 24 controller--trajectory combinations, the trajectory-wise RMSE of the direct FK--MoCap position discrepancy was $12.61 \pm 2.15$~mm (mean $\pm$ standard deviation), demonstrating that nominal kinematic reconstruction did not fully represent the measured end-effector motion. This discrepancy remained substantial under all three controller configurations. Furthermore, the MoCap-based tracking RMSE was consistently larger than the FK-based tracking RMSE across all eight trajectories, indicating that relatively small FK-based tracking errors did not imply equally small errors in the realized end-effector motion. The MoCap-measured base translation in Figure~\ref{fig:timeseries_experiment} varies by case during trajectory tracking, and its effects on EE tracking are among the error components to be compensated together with those induced by other uncertainties.

Averaged over the eight cases, the MoCap-based mean RMSEs are 20.13~mm, 18.26~mm, and 16.42~mm for the nominal controller, the w/o EE IMU policy, and the w/ EE IMU policy, respectively, while the corresponding mean RMSE gaps relative to FK are 5.25~mm, 4.05~mm, and 3.52~mm. The proposed w/ EE IMU policy therefore reduces the MoCap-based mean RMSE by 18.4\% relative to the nominal controller and by an additional 10.1\% relative to the w/o EE IMU policy. The larger additional benefit of EE inertial feedback on hardware (10.1\% versus 5.5\% in simulation) suggests that it captures motion effects induced by real-world mechanical uncertainties that are not sufficiently represented in simulation and cannot be fully inferred from joint-space proprioception. These results experimentally validate the central motivation of the proposed framework: a non-negligible component of the realized end-effector tracking error is not captured by joint-space proprioception and nominal kinematic reconstruction alone, while end-effector inertial feedback provides complementary information for compensating such propagated errors.

\section{Conclusion}

This paper demonstrated that direct end-effector inertial feedback can improve continuous trajectory-tracking accuracy on a quadruped manipulator under substantial mechanical uncertainties.
An end-effector-mounted IMU enables the learned controller to respond to realized end-effector motion after errors from base motion, contact, actuation, and kinematic uncertainty have propagated through the leg-arm system.
Unlike joint-space proprioception alone, this feedback provides information about task-space motion without requiring external sensing infrastructure during control. The two-stage sim-to-real alignment enabled hardware deployment of the inertial-feedback policy by reducing discrepancies in the arm dynamics and EE inertial observations. The real-world experiments also showed a consistent gap between FK-based and MoCap-based tracking errors, indicating that a non-negligible portion of the realized end-effector error cannot be captured by joint encoders and nominal FK alone.

Based on the mean RMSE across all eight evaluation cases, the proposed policy achieved a 30.5\% reduction relative to the nominal controller and an additional 5.5\% reduction relative to the proprioception-only residual policy in simulation. In real-world experiments, the corresponding mean MoCap-based RMSE reductions were 18.4\% and 10.1\%, respectively. The larger additional benefit of EE inertial feedback on hardware suggests that it captures motion effects induced by real-world mechanical uncertainties that are not sufficiently represented in simulation and cannot be fully inferred from joint-space proprioception. The proposed method also generally achieved the lowest tracking errors under out-of-distribution commands with velocities and/or accelerations beyond the training range, demonstrating the effectiveness of end-effector inertial feedback for accurate trajectory tracking under propagated system uncertainties.
The present study intentionally focuses on local high-accuracy manipulation without deliberate body motion or foot repositioning; extending the framework to workspace-expanding whole-body motions and higher-DOF mobile manipulators, such as humanoids, is left for future work.

\renewcommand{\IEEEbibitemsep}{-1.5pt}
\IEEEtriggeratref{21}
\bibliographystyle{IEEEtran}
\bibliography{ref}

\begin{thebibliography}{10}
\providecommand{\url}[1]{#1}
\csname url@samestyle\endcsname
\providecommand{\newblock}{\relax}
\providecommand{\bibinfo}[2]{#2}
\providecommand{\BIBentrySTDinterwordspacing}{\spaceskip=0pt\relax}
\providecommand{\BIBentryALTinterwordstretchfactor}{4}
\providecommand{\BIBentryALTinterwordspacing}{\spaceskip=\fontdimen2\font plus
\BIBentryALTinterwordstretchfactor\fontdimen3\font minus \fontdimen4\font\relax}
\providecommand{\BIBforeignlanguage}[2]{{%
\expandafter\ifx\csname l@#1\endcsname\relax
\typeout{** WARNING: IEEEtran.bst: No hyphenation pattern has been}%
\typeout{** loaded for the language `#1'. Using the pattern for}%
\typeout{** the default language instead.}%
\else
\language=\csname l@#1\endcsname
\fi
#2}}
\providecommand{\BIBdecl}{\relax}
\BIBdecl

\bibitem{decoupled1}
K.~Xu, Q.~Li, Y.~Zhao, J.~Chen, and X.~Ding, ``Accurate and efficient end-effector tracking for quadrupedal mobile manipulators through coordinated motion and decoupled control,'' \emph{Adv. Intell. Syst.}, vol.~7, no.~8, p. 2500242, 2025.

\bibitem{wholebody_RL2}
T.~Portela, A.~Cramariuc, M.~Mittal, and M.~Hutter, ``Whole-body end-effector pose tracking,'' in \emph{Proc. IEEE Int. Conf. Robot. Autom. (ICRA)}, 2025, pp. 11\,205--11\,211.

\bibitem{wholebody_RL3}
G.~Pan, Q.~Ben, Z.~Yuan, G.~Jiang, Y.~Ji, S.~Li, J.~Pang, H.~Liu, and H.~Xu, ``{RoboDuet}: Learning a cooperative policy for whole-body legged loco-manipulation,'' \emph{IEEE Robot. Autom. Lett.}, vol.~10, no.~5, pp. 4564--4571, 2025.

\bibitem{wholebody_RL4}
Z.~Fu, X.~Cheng, and D.~Pathak, ``Deep whole-body control: Learning a unified policy for manipulation and locomotion,'' in \emph{Proc. Conf. Robot Learn. (CoRL)}, 2023, pp. 138--149.

\bibitem{wholebody_RL5}
H.~Ha, Y.~Gao, Z.~Fu, J.~Tan, and S.~Song, ``{UMI} on legs: Making manipulation policies mobile with manipulation-centric whole-body controllers,'' \emph{arXiv preprint arXiv:2407.10353}, 2024.

\bibitem{wholebody_RL6}
D.~Hou, C.~Zhu, Z.~Zhang, Z.~Li, C.~Guo, and Y.~Liu, ``Efficient learning of a unified policy for whole-body manipulation and locomotion skills,'' in \emph{Proc. IEEE/RSJ Int. Conf. Intell. Robots Syst. (IROS)}, 2025, pp. 5455--5461.

\bibitem{wholebody_RL7}
N.~Fey, G.~B. Margolis, M.~Peticco, and P.~Agrawal, ``Bridging the sim-to-real gap for athletic loco-manipulation,'' \emph{arXiv preprint arXiv:2502.10894}, 2025.

\bibitem{wholebody_modelbased1}
D.~Wang, J.~Yu, S.~Wu, Z.~Li, C.~Li, R.~Xiong, S.~Qu, and Y.~Wang, ``A hierarchical {MPC} for end-effector tracking control of legged mobile manipulators,'' \emph{IEEE Trans. Autom. Sci. Eng.}, vol.~22, pp. 4855--4866, 2025.

\bibitem{wholebody_modelbased2}
M.~Risiglione, V.~Barasuol, D.~G. Caldwell, and C.~Semini, ``A whole-body controller based on a simplified template for rendering impedances in quadruped manipulators,'' in \emph{Proc. IEEE/RSJ Int. Conf. Intell. Robots Syst. (IROS)}, 2022, pp. 9620--9627.

\bibitem{wholebody_modelbased3}
Y.~Ma, F.~Farshidian, T.~Miki, J.~Lee, and M.~Hutter, ``Combining learning-based locomotion policy with model-based manipulation for legged mobile manipulators,'' \emph{IEEE Robot. Autom. Lett.}, vol.~7, no.~2, pp. 2377--2384, 2022.

\bibitem{wholebody_modelbased4}
H.~Ferrolho, V.~Ivan, W.~Merkt, I.~Havoutis, and S.~Vijayakumar, ``Roloma: Robust loco-manipulation for quadruped robots with arms,'' \emph{Auton. Robots}, vol.~47, no.~8, pp. 1463--1481, 2023.

\bibitem{wheeled_wholebody1}
K.~Jiang, Z.~Fu, J.~Guo, W.~Zhang, and H.~Chen, ``Learning whole-body loco-manipulation for omni-directional task-space pose tracking with a wheeled-quadrupedal manipulator,'' \emph{IEEE Robot. Autom. Lett.}, vol.~10, pp. 1481--1488, 2025.

\bibitem{wheeled_wholebody2}
Z.~Wang, Y.~Jia, L.~Shi, H.~Wang, H.~Zhao, X.~Li, J.~Zhou, J.~Ma, and G.~Zhou, ``Arm-constrained curriculum learning for loco-manipulation of a wheel-legged robot,'' in \emph{Proc. IEEE/RSJ Int. Conf. Intell. Robots Syst. (IROS)}, 2024, pp. 10\,770--10\,776.

\bibitem{wholebody_RL1}
T.~Portela, G.~B. Margolis, Y.~Ji, and P.~Agrawal, ``Learning force control for legged manipulation,'' in \emph{Proc. IEEE Int. Conf. Robot. Autom. (ICRA)}, 2024, pp. 15\,366--15\,372.

\bibitem{QDD1}
S.~Seok, A.~Wang, M.~Y. Chuah, D.~J. Hyun, J.~Lee, D.~M. Otten, J.~H. Lang, and S.~Kim, ``Design principles for energy-efficient legged locomotion and implementation on the {MIT} cheetah robot,'' \emph{IEEE/ASME Trans. Mechatronics}, vol.~20, no.~3, pp. 1117--1129, 2015.

\bibitem{global_feedback2}
M.~Liu, Z.~Chen, X.~Cheng, Y.~Ji, R.-Z. Qiu, R.~Yang, and X.~Wang, ``Visual whole-body control for legged loco-manipulation,'' in \emph{Proc. Conf. Robot Learn. (CoRL)}, 2025, pp. 234--257.

\bibitem{totalstation1}
D.~S. Paraforos, M.~Reutemann, G.~Sharipov, R.~Werner, and H.~W. Griepentrog, ``Total station data assessment using an industrial robotic arm for dynamic {3D} in-field positioning with sub-centimetre accuracy,'' \emph{Comput. Electron. Agric.}, vol. 136, pp. 166--175, 2017.

\bibitem{laser_tracker1}
Z.~Wang, R.~Zhang, and P.~Keogh, ``Real-time laser tracker compensation of robotic drilling and machining,'' \emph{J. Manuf. Mater. Process.}, vol.~4, no.~3, p.~79, 2020.

\bibitem{IMU1}
J.~Kindle, M.~Loetscher, A.~Alessandretti, C.~Cadena, and M.~Hutter, ``Enhancing robotic precision in construction: A modular factor graph-based framework to deflection and backlash compensation using high-accuracy accelerometers,'' \emph{IEEE Robot. Autom. Lett.}, vol.~10, no.~1, pp. 288--295, 2025.

\bibitem{IMU4}
L.~Stadelmann, T.~Sandy, A.~Thoma, and J.~Buchli, ``End-effector pose correction for versatile large-scale multi-robotic systems,'' \emph{IEEE Robot. Autom. Lett.}, vol.~4, no.~2, pp. 546--553, 2019.

\bibitem{IMU3}
T.~Cvitanic, S.~Melkote, and S.~Balakirsky, ``Improved state estimation of a robot end-effector using laser tracker and inertial sensor fusion,'' \emph{CIRP J. Manuf. Sci. Technol.}, vol.~38, pp. 51--61, 2022.

\bibitem{IMU5}
M.~H. Korayem, M.~Yousefzadeh, and S.~Kian, ``Precise end-effector pose estimation in spatial cable-driven parallel robots with elastic cables using a data fusion method,'' \emph{Measurement}, vol. 130, pp. 177--190, 2018.

\bibitem{IMU6}
T.~Sandy and J.~Buchli, ``Dynamically decoupling base and end-effector motion for mobile manipulation using visual-inertial sensing,'' in \emph{Proc. IEEE/RSJ Int. Conf. Intell. Robots Syst. (IROS)}, 2017, pp. 6299--6306.

\bibitem{IMU7}
S.~H. Hashemi and J.~Mattila, ``Hybrid controller for robot manipulators in task-space with visual-inertial feedback,'' \emph{IEEE Trans. Control Syst. Technol.}, vol.~34, no.~2, pp. 737--746, 2026.

\bibitem{Residual_RL1}
T.~Johannink, S.~Bahl, A.~Nair, J.~Luo, A.~Kumar, M.~Loskyll, J.~A. Ojea, E.~Solowjow, and S.~Levine, ``Residual reinforcement learning for robot control,'' in \emph{Proc. IEEE Int. Conf. Robot. Autom. (ICRA)}, 2019, pp. 6023--6029.

\bibitem{isaaclab}
M.~Mittal, P.~Roth, J.~Tigue, A.~Richard, O.~Zhang, P.~Du, A.~Serrano-Mu{\~n}oz, X.~Yao, R.~Zurbr{\"u}gg, N.~Rudin \emph{et~al.}, ``{Isaac Lab}: A {GPU}-accelerated simulation framework for multi-modal robot learning,'' \emph{arXiv preprint arXiv:2511.04831}, 2025.

\bibitem{PPO}
J.~Schulman, F.~Wolski, P.~Dhariwal, A.~Radford, and O.~Klimov, ``Proximal policy optimization algorithms,'' \emph{arXiv preprint arXiv:1707.06347}, 2017.

\end{thebibliography}

\end{document}